\documentclass[letterpaper, 10 pt, conference]{ieeeconf}  % Comment this line out if you need a4paper

\IEEEoverridecommandlockouts                              % This command is only needed if 
\usepackage{graphicx}
\usepackage{epsfig}
\usepackage{epstopdf}
\usepackage{times}
\usepackage{newtxtext}

\usepackage[T1]{fontenc}  % 让 LaTeX 处理现代字体
\usepackage{lmodern}      % 使用 Latin Modern 字体
\usepackage{graphicx}

\usepackage{amsmath}    % 数学公式
\usepackage{amssymb}    % 数学符号
\usepackage{cases}      % 多行公式
\usepackage{algorithmicx} % 算法伪代码
\usepackage{algorithm}       % 算法浮动体支持
\usepackage{algpseudocode}  % 算法伪代码样式
\usepackage{autobreak}

\usepackage{tabularx}   % 弹性表格环境
\usepackage{booktabs}  %% 三线表
\usepackage{multirow}  %% 合并单元格
\usepackage[table,xcdraw]{xcolor}  % 用于设置表格单元格的颜色

\usepackage{hyperref}
\usepackage{cleveref}
\usepackage{subcaption}
\usepackage{float}

\usepackage{xcolor}
\usepackage{soul}

\usepackage{fancyhdr}

\usepackage{titlesec}

\title{\LARGE \bf
HIF: Height Interval Filtering for Efficient Dynamic Points Removal
}

\author{Shufang Zhang$^{1}$, Tao Jiang$^{1}$, Jiazheng Wu$^{1}$, Ziyu Meng$^{2}$, Ziyang Zhang$^{1}$ and Shan An$^{1}$% <-this % stops a space
\thanks{$^{1}$Shufang Zhang, Tao Jiang, Jiazheng Wu, Ziyang Zhang and Shan An are with the School of Electrical and Information Engineering, Tianjin University, Tianjin 300072, China. {\tt\small\{shufangzhang, jtao, wujiazheng, zhang\_ziyang, anshan\}@tju.edu.cn}} %
\thanks{$^{2}$Ziyu Meng is with the School of Control Science and Engineering, Shandong University, Jinan 250061, China. mziyu@mail.sdu.edu.cn}%
}

\fancypagestyle{firstpage}{
    \fancyhf{} % 清除默认页眉页脚
    \fancyhead[L]{\small This work has been submitted to the IEEE for possible publication. Copyright may be transferred without notice, after which this version may no longer be accessible.}
}

\begin{document}

% 应用第一页的页眉样式
\maketitle
\thispagestyle{firstpage}

\maketitle
% \thispagestyle{empty}
% \pagestyle{empty}

%%%%%%%%%%%%%%%%%%%%%%%%%%%%%%%%%%%%%%%%%%%%%%%%%%%%%%%%%%%%%%%%%%%%%%%%%%%%%%%%
\begin{abstract}
        3D point cloud mapping plays a essential role in localization and autonomous navigation. 
        However, dynamic objects often leave residual traces during the map construction process, which undermine the performance of subsequent tasks. 
        Therefore, dynamic object removal has become a critical challenge in point cloud based map construction within dynamic scenarios. Existing approaches, however, often incur significant computational overhead, making it difficult to meet the real-time processing requirements. To address this issue, we introduce the Height Interval Filtering (HIF) method. 
        This approach constructs pillar-based height interval representations to probabilistically model the vertical dimension, with interval probabilities updated through Bayesian inference. It ensures real-time performance while achieving high accuracy and improving robustness in complex environments.
        % Additionally, we propose a low-height preservation strategy that enhances the detection of unknown spaces, mitigating misclassification in occluded regions. 
        Additionally, we propose a low-height preservation strategy that enhances the detection of unknown spaces, reducing misclassification in areas blocked by obstacles~(occluded regions).
        % Evaluation on public datasets shows that HIF achieves optimal real-time performance, ensuring dynamic point removal quality while using minimal memory resources. 
        Experiments on public datasets demonstrate that HIF delivers a $7.7 \times$ improvement in time efficiency with comparable accuracy to existing SOTA methods. The code will be publicly available.

\end{abstract}

\section{Introduction}
As a fundamental type of observations utilized in Simultaneous Localization and Mapping (SLAM) systems, 3D point cloud has demonstrate significant potential to impact a diverse range of applications \cite{shan2020lio,xu2022fast}. In most cases, point cloud maps are constructed through a mapping process in LiDAR-equipped SLAM systems. However, complex urban environments often feature a multitude of dynamic objects, including moving vehicles, cyclists, and pedestrians, which leave residual artifacts (or ghost traces), during the mapping process, thereby significantly affecting the construction quality \cite{yoon2019mapless,pagad2020robust}. With the rapid development of point cloud mapping, a common technique is to remove dynamic objects during the mapping process.

\begin{figure}[thpb]
        \centering
        \includegraphics[width=0.95\columnwidth]{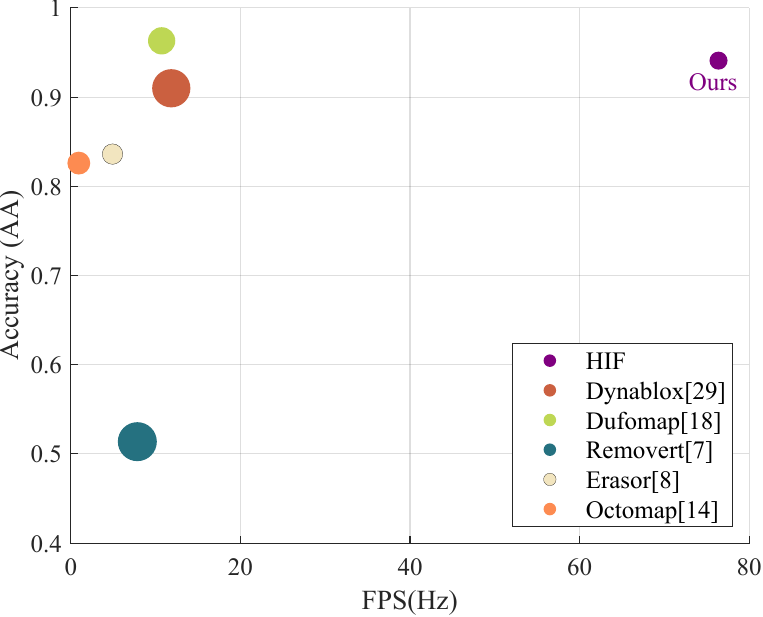}
        \caption{Comparison of FPS, accuracy and memory usage of different algorithms. The size of the points representing the algorithm's runtime memory usage.}
        %各算法运行速度与准确度比较。我们的算法使用proposed（应该加粗）标出点的大小表示算法运行时最大内存使用量。
\label{fig:runtime_accuracy}
\end{figure}

 % To circumvent this problem, the removal of dynamic objects has garnered attention from researchers.
 
 % map 前需要修饰

% 1.传统方法需要把点放到voxel里，检索成本高 2.前后帧扫描计算
% strength :1. 二维检索快，用哈希转为一维检索. 2. 不需要点对点操作
%。  3. 不是全局信息比对，检索到的才比对

% 有三维哈希检索，但我们是第一个做到二维的(trick)

% !!!!! Fig 2没有在文中被引用

%%地面点误杀用词：incorrectly seen as dynamic

%点云地图被广泛应用于导航、定位以及同时定位与地图构建（SLAM）系统。通常，点云地图是由配备激光雷达（LiDAR）的SLAM系统通过建图过程生成的。然而，在复杂的城市环境中，存在大量动态物体，如移动车辆、行人和自行车等。这些动态物体在建图过程中会留下残迹，对下游任务产生严重影响。因此，动态物体去除任务逐渐受到研究者的关注。

Traditionally, dynamic point removal is often implemented using ray casting.
% 批评 ray-casting
However, this approach relies heavily on computationally intensive retrieval processes, involving point-by-point operations that require traversing each voxel, making it challenging to meet real-time requirements. Moreover, due to the small pitch angle between the rays and the distant ground surface, this method is prone to misclassifying ground points, leading to incorrect removals~\cite{fu2022mapcleaner, wu2024otd}.
% 只指出了问题，没有说明原因————>原因是远处的地面光线与地面的角度较小angle between the rays and the ground line becomes very small（观点来源fu2022mapcleaner，wu2024otd）
% 前面是ray-casting的问题，1.慢 2.ground points error

% recent study是怎么做的，这里补充两句描述，后处理找不同。recent study需要一个概括性描述。-->recent sutdy将动态物体检测任务转换为检测帧与帧之间的占用差异。wu2024otd
Recent studies \cite{kim2020remove,wu2024otd,lim2021erasor,zhang2024erasor++,jia2024beautymap,fu2022mapcleaner} have formulated dynamic object removal as detecting occupancy differences between consecutive frames. However, most of these approaches operate as post-processing methods, requiring all scans to be treated as global priors to ensure accuracy, which prevents real-time execution. Additionally, some methods employ egocentric map representations that lack global consistency, necessitating extra transformations from global map points to the local frame \cite{jia2024beautymap}, thereby introducing significant computational overhead. Moreover, many of these approaches rely on ground segmentation modules, further increasing both computational complexity and system intricacy.

% 总结共性问题，总结的问题需要和你的方法对应，写你的方法解决了的问题
%Nevertheless, most of these methods are \hl{}, .
% computational-intensive and lack support for online execution
% 原理上的问题->计算量大->速度慢->real-time, application 差
%These approaches typically require dedicated ground segmentation modules to preserve ground points, introducing additional computational overhead and system complexity, thereby limiting their applicability.

% \begin{figure}[thpb]
%     \centering
%     \includegraphics[width=0.8\linewidth]{open.pdf}
%     \caption{Illustration of the HIF algorithm. The top image displays the original data from the Semantic KITTI dataset, while the bottom image presents the processed result using the HIF algorithm.}
%     \label{fig:opening}
% \end{figure}

% 核心贡献 总结 全局一致的pillar
In this paper, we propose a dynamic point cloud removal method, termed Height Interval Filtering (HIF). This method partitions the space into globally consistent pillars and employs a \textbf{hash-mixed indexing} operation to reduce retrieval dimensions and collision rates, thereby significantly enhancing retrieval efficiency. Simultaneously, we establish height intervals to discretize the vertical space and assign probability values to them, which are then refined through Bayesian filtering.
% 补充，加速
Our method eliminates the reliance on ray-tracing and ground point segmentation modules, significantly accelerating computation and enabling efficient real-time operation.
% complex scenario 具体一下，分析场景的什么特点造成了前面的问题。特点和问题需要对应。
% 我们的方法需要能针对这个问题解决，我们方法的特点为什么解决这个问题（稍长的解释，三四句）
In real-world scenarios, complex occlusions in the environment often hinder the performance of ray-casting-based dynamic object removal algorithms, leading to decreased accuracy and efficiency. To address this, we introduce a low-height preservation strategy, which exploits the spatial structure of pillars to identify unknown spaces in the scan, thereby improving robustness in complex environments.
To validate the effectiveness of our method, we conducted experiments on the KITTI \cite{geiger2013vision} datasets with annotation labels and poses from SemanticKITTI \cite{behley2019semantickitti} and Argoverse 2 \cite{wilson2023argoverse} datasets. As shown in Fig.~\ref{fig:runtime_accuracy}, our method achieves a 6-7× increase in FPS compared to recent methods, while maintaining high accuracy in dynamic object removal.
% The experimental results, as shown in Fig.~\ref{fig:runtime_accuracy}, demonstrate that, compared to recent studies, our method achieves a 6-7 times improvement in FPS while also performing excellently in dynamic object removal accuracy.

%%在本文中，我们提出了一种在线动态点云去除方法，即高度区间滤波（HIF）。我们将空间划分为全局一致的pillarpillar，并构建高度区间对垂直方向进行概率建模，利用贝叶斯滤波更新区间概率。我们的方法克服了对射线追踪和地面点分割模块的依赖，实现在线实时运行。我们还提出了一种低高度保持策略，能够分析每帧扫描中的未知空间，从而在复杂场景下提升方法的表现。为了验证我们方法的有效性，我们在KITTI和AV2数据集上进行了实验。实验结果显示如Fig.~\ref{fig:runtime_accuracy}，相比近期的研究我们的方法取得了6~7倍的FPS提升，在动态物体去除的准确性同样表现出色。

Briefly, we identify our contributions as follows

\begin{itemize}
        % \item We propose the Height Interval Filtering (HIF) method, which uses height intervals for probabilistic modeling of the vertical direction, enabling efficient dynamic point cloud removal.
        \item An efficient Height Interval Filtering (HIF) approach that assigns probabilities to height intervals to model vertical structures, ensuring both efficiency and accuracy in dynamic point cloud removal.
        % 思路
        \item A low-height preservation strategy to reduce misclassification in occluded regions by identifying unknown spaces in each scan.
        % effectively identifying unknown spaces in each scan. effectively需要拓展一句
        % 具体实现
        % \item We show the capability of our proposed HIF to remove dynamic objects with superior runtime performance while maintaining high accuary on KITTI and AV2 datasets.
        \item Evaluated on the KITTI and AV2 datasets, the proposed HIF algorithm achieves exceptional computational efficiency while preserving high accuracy.
\end{itemize}

%我们提出HIF方法，通过构建高度区间对垂直方向进行概率建模，使用简洁的框架实现了高效的动态点云去除。
%我们提出了低高度保持策略，高效的感知单帧扫描中unknown的空间减少遮挡区域误分类。
%我们在KITTI以及AV2数据集上验证了HIF方法的有效性，证明了其在保证动态物体去除效果的同时，实现最快的运行速度。

%The structure of the paper is as follows: In Section 2, we review existing methods for dynamic point cloud removal. In Section 3, we provide a detailed explanation of our height interval filtering method. In Section 4, we present our experimental results, comparing them with existing methods and conducting ablation experiments. Finally, we conclude the paper and provide an outlook in Section 5.

%%我们文章安排如下：在第二章中，我们介绍了现有的动态点云去除方法。在第三章中，将详细介绍我们的高度区间滤波方法。在第四章中，我们将展示我们的实验结果，与现有的方法进行比较以及消融实验。最后，我们在第五章中对全文进行总结和展望。

%wu2024observation中的简介中提到，离线方法指的是单帧与先验地图（拥有所有时刻的信息），在线的方式指的每刻比对的只有当前帧之前的信息

\section{Related Works}

% In this chapter, we review the existing approaches for dynamic point cloud removal, categorizing them into three main types: freeSpace-based Removal methods, difference-based methods and learning-based methods.

\subsection{FreeSpace-based Removal methods}

\begin{figure}[htp]
    \centering
    \includegraphics[width=0.95\columnwidth]{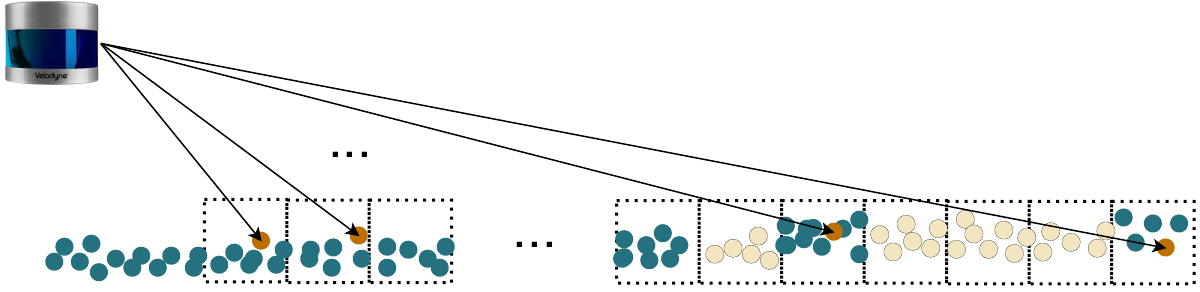}
    \caption{Ground point misclassification in ray-tracing. 
    Points in \textcolor[rgb]{0.941, 0.635, 0.418}{(orange)} represent the current scan, while points in 
    \textcolor[rgb]{0.9, 0.7, 0.2}{(dark yellow)} indicate ground points mistakenly classified as dynamic.}
    \label{raycast}
\end{figure}

FreeSpace-based removal methods represent point clouds as voxels and update occupancy states using per-frame scan results based on ray-tracing principles \cite{hornung2013octomap, schauer2018peopleremover}. However, sole reliance on ray-tracing leads to significant misclassification issues, particularly affecting ground points, as shown in Fig.~\ref{raycast}. To address this, some methods incorporate post-processing techniques to recover ground points after ray-tracing, effectively reducing misclassification errors \cite{zhang2023dynamic, arora2023static}. Additionally, improvements in trace efficiency and the introduction of more robust discriminative mechanisms have further enhanced both computational efficiency and accuracy \cite{duberg2024dufomap, yan2023rh}. Despite these advancements, computationally expensive 3D retrieval remains a major bottleneck for real-time performance. Our approach mitigates this by partitioning the space into pillars and transforming 3D retrieval into a more efficient 1D lookup using a hash-mixed operation, significantly reducing computational complexity.

%%地面点误杀的原因的另一种说法severe misclassification caused by incidence angle ambiguity（来源\cite{yan2023rh}，观点来自\cite{lim2021erasor}）

%%Occupancy voxel-based methods的方法将点云转换为体素表示并且根据射线追踪的原理使用每帧扫描结果来更新体素占据状态\cite{hornung2013octomap} \cite{schauer2018peopleremover}。但是仅仅使用光线追踪会带来严重的误杀问题。\cite{zhang2023dynamic}和\cite{arora2023static}，通过在光线追踪后恢复地面点来减少地面点的误杀。 \cite{duberg2024dufomap} \cite{yan2023rh} 上优化了trace效率同时采取更加科学的判别机制实现了计算效率与性能的双重提升。尽管现有的方法不断进行优化，但是计算密集的三维检索仍然对方法实时运行带来巨大的挑战。在我们的工作中，我们通过将空间划分为pillars并通过哈希混合操作将三维检索复杂度压缩至一维检索，极大程度降低了检索复杂度。

\subsection{Difference-based methods}

Difference-based methods detect dynamic objects by analyzing visibility or disparity changes. One approach involves converting maps into depth images and comparing depth differences between frames and a global map, as demonstrated in \cite{kim2020remove, fu2022mapcleaner}. Other methods prioritize vertical information, detecting changes along the vertical axis to identify dynamic objects \cite{lim2021erasor, zhang2024erasor++, jia2024beautymap, wu2024otd}. Some techniques employ vertical layering, which can lead to rigid height stratification issues. Most difference-based approaches operate as post-processing techniques, requiring all scans as global priors to ensure accuracy, thereby limiting real-time applicability. Additionally, these methods heavily depend on ground segmentation, increasing computational overhead and system complexity. In contrast, our approach models vertical distributions by constructing flexible height intervals, eliminating reliance on ground segmentation and mitigating rigid height stratification issues.

%%基于差异的方法通过可见性或者差别检测的方式，发现动态物体。\cite{kim2020remove}\cite{fu2022mapcleaner}通过将地图转换为深度图比较单帧与全局地图的深度图差异。\cite{lim2021erasor}\cite{erasor++}\cite{jia2024beautymap}\cite{wu2024otd}更加关注垂直方向信息，检测垂直方向的变化来发现动态物体，部分方法采用垂直分层的思想，这存在高度分层僵化的潜在问题。多数基于差异的方法都是后处理的方式，需要提前获取所有扫描作为全局先验以保证真确率，限制了这些方法的应用场景。同时地面点在这些方法中通常扮演了重要的角色，因此这些方法十分依赖地面分割模块，这在一定程度上增加了计算量以及系统复杂度。相比之下我们的方法采用构建灵活的高度区间对垂直方向进行概率建模，避免了对地面分割模块的依赖同时不存在高度分层僵化的问题。

\begin{figure*}[htpb]
    \centering
    \vspace{0.2cm}
    \includegraphics[width=\textwidth]{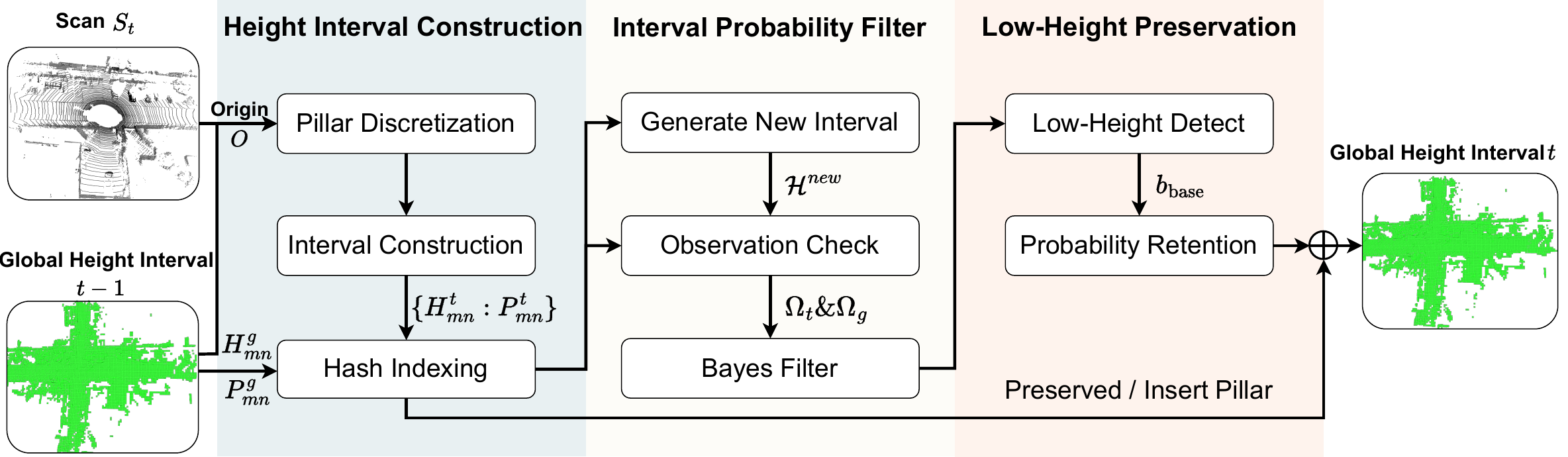}
    \caption{Framework of the HIF algorithm. It consists of three main modules: \textit{Height Interval Construction}, \textit{Interval Probability Filtering}, and \textit{Low-Height Preservation}. The algorithm processes an input scan $S_t$ at time $t$.}
    \label{Framework}
\end{figure*}

\subsection{learning-based methods}

Learning-based methods have shown strong capabilities across various domains. In dynamic object removal, the task is often reformulated as a point cloud segmentation problem \cite{milioto2019rangenet++, cortinhal2020salsanext, mersch2022receding}. These approaches transform point clouds into range images to achieve a dense and compact representation \cite{milioto2019rangenet++, cortinhal2020salsanext}. More recent methods introduce residual images to incorporate dynamic information, improving segmentation performance \cite{chen2021moving, sun2022efficient, li2023efficient, xie2023real, cheng2024mf}. However, learning-based approaches require extensive training data and suffer from data imbalance issues \cite{zhang2022not}, while generalization and robustness remain significant challenges \cite{duberg2024dufomap}. Furthermore, their high computational cost limits practical deployment.

%学习的方法在很多领域展现出强大的能力，对于动态物体去除任务通常被转换成点云分割任务\cite{milioto2019rangenet++}\cite{cortinhal2020salsanext}\cite{mersch2022receding}.将点云转换成range image从而实现点云的密集且紧凑表示\cite{milioto2019rangenet++} \cite{cortinhal2020salsanext}，最近的工作通过引入Residual Images来增加动态信息以提升动态物体分割的性能\cite{chen2021moving} \cite{sun2022efficient} \cite{li2023efficient} \cite{xie2023real}\cite{cheng2024mf}。这个文章指出点云的学习方法对大量训练的需求且存在数据不平衡的问题\cite{zhang2022not}，泛化性和鲁棒性存在挑战\cite{duberg2024dufomap}。且基于学习的方法通常需要大量计算资源限制其应用场景。

\section{Methodology}

We propose Height Interval Filtering (HIF) for dynamic point removal, as illustrated in Fig.~\ref{Framework}. HIF consists of three key steps: height interval construction (Section~\ref{height_interval_construction}), interval probability filtering (Section~\ref{interval_probability_filtering}), and low-height preservation (Section~\ref{LHP_S}). Upon acquiring a new point cloud scan, it is transformed into a height interval representation using a globally consistent partitioning scheme, with global height intervals updated via Bayesian filtering. The proposed low-height preservation strategy identifies unknown space in the scan, improving retention of static structures in highly occluded environments. Finally, dynamic points are removed by preserving those within height intervals with high static probabilities.

%%我们提出了一种基于高度区间的动态点云去除方法，如图~\ref{Framework}所示。该方法主要包含三个步骤：高度区间构造（Section III-A）、区间概率滤波（Section III-B）以及低高度保留（Section III-C）。每当获取新的点云扫描，将其转换为全局统一划分方式的高度区间表示，进而基于贝叶斯滤波的原理对全局高度区间进行更新。此外，我们提出的低高度保留策略能够有效感知扫描中的未知空间，提高复杂场景下的表现。最终，当扫描序列处理完成后，通过保留静态概率较高的高度区间中的点，即可完成动态点去除任务。

\subsection{Height Interval Construction}
\label{height_interval_construction}

Recent studies \cite{lim2021erasor, zhang2024erasor++, jia2024beautymap, wu2024otd} have demonstrated that analyzing vertical information effectively aids in dynamic object identification. Inspired by these works, we adopt a pillar-based representation to capture vertical structures. Within each pillar, we construct height intervals to model spatial occupancy probabilities. To ensure a globally consistent pillar partitioning, we define a fixed origin point $O$ on the XY plane, preventing misalignment between new scans and the global height intervals.

%%近期的研究，\cite{lim2021erasor}、\cite{zhang2024erasor++}、\cite{jia2024beautymap}和\cite{wu2024otd}，表明比对垂直维度的信息可以有效识别动态物体。受此启发，我们采用基于立柱（pillar）的数据表示方式，关注垂直方向信息，并在每个立柱内通过构建高度区间来维护空间的占用概率。此外，我们在XY平面上设置了一个原点 O，以确保新扫描与全局高度区间有着统一的pillar划分。

In our approach, we represent the environment using a \textbf{2D pillar-based grid} where each pillar has a fixed size of $\Delta x \times \Delta y$ in the XY plane. These pillars are used to discretize the space horizontally, while vertical information is modeled by constructing height intervals within each pillar to maintain spatial occupancy probabilities.

At time $t$, the point cloud scan is denoted as $S_t = \{ \mathbf{p}_i = (x_i, y_i, z_i) \mid i = 1, 2, ..., N \}$, where $N$ represents the total number of points in the scan. Each point $\mathbf{p}_i \in S_t$ is assigned to a corresponding pillar $P_{mn}$ by computing its offset $(m, n)$ relative to the global origin $O = (x_o, y_o)$. The offset is determined as follows:
\begin{equation}
        m = \left\lfloor \frac{x_i - x_o}{\Delta x} \right\rfloor, \quad n = \left\lfloor \frac{y_i - y_o}{\Delta y} \right\rfloor
        \label{eq:pillar_index}
\end{equation}

\begin{figure}[thpb]
        \centering
        \vspace{0.1cm}
        \includegraphics[width=0.95\columnwidth]{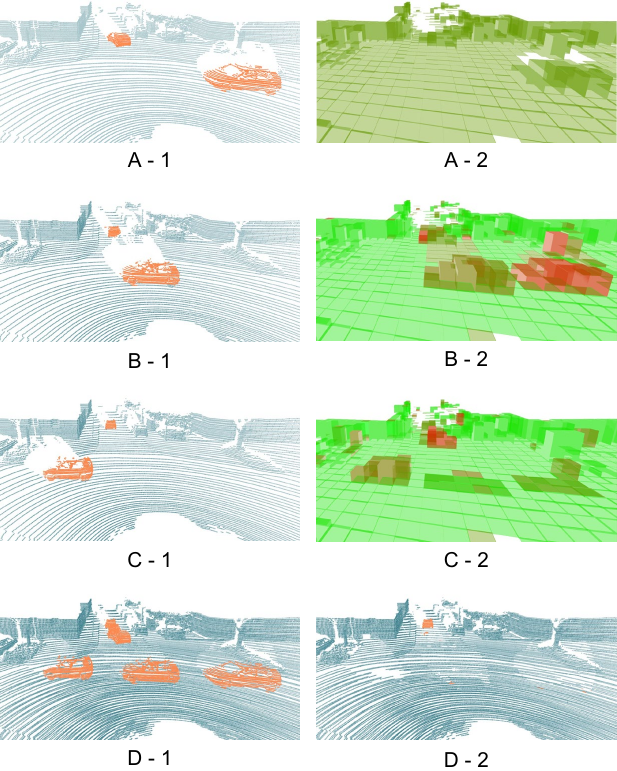}  % 替换为您的PDF图片路径
        \caption{Schematic of the \textbf{HIF} algorithm. It illustrates three point cloud scans from the same scene, labeled as A-1, B-1, and C-1. \textbf{A-2} shows the height interval construction based on \textbf{A-1}. \textbf{B-2} and \textbf{C-2} demonstrate the updating process of height intervals using new point cloud scans, where \textcolor[rgb]{0.0, 0.6, 0.0}{green} represents intervals with high static probability, and \textcolor[rgb]{0.8, 0.0, 0.0}{red} represents intervals with low static probability. \textbf{D-1} presents the combined scans from all three frames, while \textbf{D-2} shows the result of filtering all points using the height intervals from \textbf{C-2}. Note that the parameter values in this figure have been adjusted slightly to better illustrate the algorithm's effectiveness.}
        %HIF算法原理示意图展示了我们在同一场景下选取的三帧点云扫描数据，如A-1、B-1和C-1。A-2是由A-1进行高度区间构建后的结果，B-2和C-2则展示了根据新的点云扫描对高度区间进行更新的过程，其中绿色表示静态概率较高的区间，红色表示静态概率较低的区间。D-1展示了三帧扫描数据的叠加结果，D-2则是使用C-2的高度区间对所有点进行过滤后的结果。需要注意的是，图中所用的参数值与实际算法略有不同，目的是为了更好地展示算法的有效性。
        \label{HIF_algorithm}
\end{figure}

%%在我们的方法中，我们设定每个pillar的大小为$\Delta x \times \Delta y$。对于时刻$t$的扫描$S_t$中的每个点$\mathbf{p}i = (x_i, y_i, z_i) \in S_t$，我们通过计算该点在XY平面上相对于全局原点$O = (x_o, y_o)$的偏移量$(m,n)$，来确定其所属的pillar$ P{mn} $。具体来说，偏移量$(m,n)$通过以下公式计算：

%%这个划分方式确保每个pillar $ P{mn} $的boundaries满足：
This scheme ensures the pillar $P{mn}$ boundaries align with:
\begin{equation}
        P_{mn} = \left\{ \mathbf{p}_i \, \bigg| \, 
        \begin{aligned}
        x_o + m\Delta x \leq &x_i < x_o + (m + 1)\Delta x \\
        y_o + n\Delta y \leq &y_i < y_o + (n + 1)\Delta y
        \end{aligned} \right\}
        \label{eq:pmn_v2}
\end{equation}

To enable efficient pillar retrieval, we employ a hash-mixed operation to generate unique pillar indices:
% \begin{equation}
% \small
%         H_{mn} = h(m) \oplus \left( h(n) + \mathrm{C} + (h(m) \ll 6) + (h(m) \gg 2) \right)
% \end{equation}

\begin{equation}
    H_{mn} = h(m) \oplus ( h(n) + \mathrm{C} + (h(m) \ll 6) + (h(m) \gg 2) )
\end{equation}

where $h(\cdot)$ represents a standard integer hash function, $\mathrm{C}$ is a constant derived from the golden ratio, commonly used in hash functions to improve distribution uniformity. The operators $\ll$ and $\gg$ denote bitwise left and right shifts, respectively. 

This hashing strategy ensures an injective mapping, minimizing the risk of collisions. Furthermore, to optimize storage efficiency, we store only non-empty pillars and construct an index for fast access.

%为了实现高效的柱体检索，我们采用了哈希混合操作：
%其中，$h(\cdot)$ 表示标准整数哈希函数，$\mathrm{C}$ 是黄金比例常数，$\ll$ 和 $\gg$ 表示位移操作。该哈希策略保证了单射映射，减少了碰撞的风险。此外，我们仅存储非空的柱体，并构建索引以进一步提高存储效率。

Next, our approach dynamically merges height values within each pillar into adaptive height intervals, enabling finer vertical representation without requiring ground segmentation, as illustrated in Fig.~\ref{HIF_algorithm}~A-1 and Fig.~\ref{HIF_algorithm}~A-2. This design addresses the limitations of existing methods, which suffer from rigid vertical partitioning. The R-POD method in \cite{lim2021erasor} and the approaches in \cite{zhang2024erasor++, jia2024beautymap} rely on predefined height divisions with binary or symbolic encoding, making them less adaptable to elevation changes and occlusions. Additionally, these methods require auxiliary ground segmentation and lack probabilistic modeling, leading to potential misclassifications in complex scenes.

%现有方法存在垂直划分僵化的问题：\cite{lim2021erasor}中的R-POD方法将空间划分为bins，并使用伪占据符号（pseudo occupancy）记录最高点和最低点之间的偏差。 \cite{zhang2024erasor++} 和 \cite{jia2024beautymap}则采用垂直方向等距划分，并通过二进制编码描述垂直占据状态。这些空间划分方式能够高效实现空间状态检测，但需要辅助地面分割且缺乏概率建模。与这些方法不同，我们的方法提取每个pillar内部点的高度，并将这些高度合并成高度区间，如图Figure \ref{HIF_algorithm} Parts A-1和A-2所示。

%在记录每个高度区间占用概率的同时，我们为每个柱体设立一个$p_{\text{empty}}$代表空白区域中得占据的概率，因此每个$P_{mn}$维护：
We define $p_{\text{empty}}$ to maintain the occupancy probability of the empty space in each pillar, which provides reasonable initialization for the observed height interval in new scans. Each pillar \( P_{mn} \) is structured as follows:
\begin{equation}
    P_{mn} = \left( p_{\text{empty}}, \, \mathcal{H}_{mn} = \left\{ h_k = (b_k, t_k, p_k) \right\}_{k=1}^{K} \right)
    \label{eq:pillar_definition}
\end{equation}
where the set $\mathcal{H}_{mn}$ represents the collection of height intervals within a pillar, containing $K$ discrete height intervals $h_k$. Each interval $h_k$ is defined by three parameters: $b_k$ and $t_k$, which denote the minimum and maximum height values within the interval, respectively, and $p_k$, which represents the occupancy probability of the interval.

We represent each pillar using an index \( H_{mn} \) and an associated storage variable \( P_{mn} \), forming a key-value pair \( H_{mn} : P_{mn} \). To maintain a global mapping of pillars, we define a dictionary \( \mathcal{G} = \{ H_{mn}^g : P_{mn}^g \} \), where \( H_{mn}^g \) denotes the spatial index of a pillar in the global map, and \( P_{mn}^g \) stores its corresponding height intervals and occupancy probabilities. Upon receiving a new scan \( S_t \) at time \( t \), we perform the Height Interval Construction operation to generate a local pillar set \( \mathcal{L}^t = \{ H_{mn}^t : P_{mn}^t \} \), where \( H_{mn}^t \) represents the spatial index of a pillar in the current scan, and \( P_{mn}^t \) contains its associated height intervals and occupancy probabilities.

For each newly generated local pillar pair $H_{mn}^t : P_{mn}^t$ in $\mathcal{L}^t$, we first check whether a corresponding global pillar $H_{mn}^g$ exists in $\mathcal{G}$. If a match is found, we update $P_{mn}^g$ using $P_{mn}^t$ following the probability filtering operations described in Sections~\ref{interval_probability_filtering} and \ref{LHP_S}. If no match is found, the new key-value pair $H_{mn}^t : P_{mn}^t$ is directly inserted into $\mathcal{G}$.

%我们维护一个全局柱体字典$\mathcal{G} = { H_{mn}^g : P_{mn}^g }$。对于由新扫描$S_t$生成的局部柱体$\mathcal{L}^t = {H_{mn}^t : P_{mn}^t }$，我们首先通过检索匹配$H_{mn}^t$与$H_{mn}^g$，然后对匹配的$P_{mn}^t$与$P_{mn}^g$执行Sec.~III-B和Sec.~III-C中的概率滤波更新操作。对于无法匹配的$H_{mn}^t : P_{mn}^t$，我们将其直接插入到全局字典$\mathcal{G}$中。

% \begin{algorithm}
%         \caption{Global Map Update Process}
%         \label{alg:map_update}
%         \begin{algorithmic}[1]
%         \State $\mathbf{Input}$: $\mathcal{G}$, $\mathcal{L}_t$
%         \For{$P_{mn}^t \in \mathcal{L}_t$}
%             \If{$H_{mn}^t \in \mathcal{G}$} 
%                 \State Retrieve and match $H_{mn}^t$ with $H_{mn}^g$ in $\mathcal{G}$
%                 \State Perform probability filtering update on $P_{mn}^t$ and $P_{mn}^g$ as described in Sec.~III-B and Sec.~III-C
%             \Else
%                 \State $\mathcal{G} \leftarrow \mathcal{G} \cup \{ H_{mn}^t : P_{mn}^t \}$
%             \EndIf
%         \EndFor
%         \end{algorithmic}
%     \end{algorithm}

\subsection{Interval Probability Filtering}
\label{interval_probability_filtering}
%我们采用二元贝叶斯滤波器更新高度区间的静态概率。观测模型参数由当前扫描状态决定：
We employ a binary Bayes filter to update the occupancy probability of height interval. The observation model parameters are determined by the height interval's observation status in the new scan $S_t$ and global mapping of pillars:
\begin{equation}
        \mathrm{Bayesfilter}(p, P_S, P_D) = \frac{P_S \cdot p}{P_S \cdot p + P_D \cdot (1 - p)}
        \label{eq:bayes_func}
        \end{equation}

\vspace{0.1cm}
\begin{equation}
        \begin{aligned}
        P_S = \begin{cases}
        \alpha & \text{if interval observed,} \\
        1-\alpha & \text{otherwise.}
        \end{cases} \\
        P_D = \begin{cases}
        \beta & \text{if interval observed,} \\
        1-\beta & \text{otherwise.}
        \end{cases}
        \end{aligned}
        \label{eq:observation_model}
        \end{equation}
where $\alpha$ and $\beta$ are empirically determined confidence parameters ($\alpha > 0.5$, $\beta < 0.5$).

When $P_{mn}^g$ is successfully matched with $P_{mn}^t$ in the new scan, it confirms that $P_{mn}^g$ has been observed. Consequently, we update the occupancy probability of empty space within $P_{mn}^g$ using Eq.~\ref{eq:empty_update}, thereby reducing the likelihood of it being occupied:
\begin{equation}
        p_{\text{empty}} \leftarrow \mathrm{Bayesfilter}(p_{\text{empty}}, 1-\alpha, 1-\beta)
        \label{eq:empty_update}
        \end{equation}
        
%当$P_{mn}^t$与$P_{mn}^g$实现哈希匹配后，意味着$P_{mn}^g$在新的扫描中被观测到，我们使用以下公式更新,这样可以降低$P_{mn}^g$中空白空间存在物体的概率。

We extract all height interval endpoints from $P_{mn}^t$ and $P_{mn}^g$ to construct the set of interval endpoints. Specifically, we extract $b_i^t$ and $t_i^t$ from the height intervals in $P_{mn}^t$, where $b_i^t$ and $t_i^t$ denote the minimum and maximum height values of the $i$-th interval $h_i^t$ in $\mathcal{H}_{mn}^t$, respectively. Similarly, we take $b_j^g$ and $t_j^g$ from all height intervals in $P_{mn}^g$, where $b_j^g$ and $t_j^g$ represent the minimum and maximum height values of the $j$-th interval $h_g^t$ in $\mathcal{H}_{mn}^g$, respectively. The resulting interval endpoint set is defined as:
  
\begin{equation}
    \mathcal{E} = \left\{ b_i^t, t_i^t \mid i=1,...,I \right\} \cup \left\{ b_j^g, t_j^g \mid j=1,...,J \right\}
    \label{eq:endpoint_union}
\end{equation}

%%我们启动高度区间融合流程。令$\mathcal{E}$表示来自两个pillar的所有端点集合： 
%%将端点排序为有序序列$\mathcal{E}^{sorted} = \{ e_1, e_2,...,e_K \}$（满足$e_k < e_{k+1}$），通过相邻端点生成新高度区间：

After sorting these endpoints into an ordered sequence $\mathcal{E}^{sorted} = \{ e_1, e_2,...,e_K \}$ where $e_k < e_{k+1}$, we generate new height intervals through pairwise combination:
\begin{equation}
        \mathcal{H}^{new} = \left\{ h_k^{new} = (e_k, e_{k+1}, p_k^{new}) \mid k=1,...,K-1 \right\}
\label{eq:new_intervals}
\end{equation}

%%对每个新区间$h_k^{new} \in \mathcal{H}^{new}$，通过重叠分析建立时域对应关系：

For each new interval \( h_k^{\text{new}} \in \mathcal{H}^{\text{new}} \), we determine its temporal correspondence by analyzing overlaps with existing height intervals. Specifically, we define \(\Omega_t\) and \(\Omega_g\) to represent the observation states of the new height interval within \(\mathcal{L}^t\) and \(\mathcal{G}\), respectively:

\begin{equation}
        \begin{aligned}
        &\Omega_t = \{ h_i^t \mid h_i^t \cap h_k^{new} \neq \emptyset \} \\
        &\Omega_g = \{ h_j^g \mid h_j^g \cap h_k^{new} \neq \emptyset \}
        \end{aligned}
\label{eq:overlap_sets}
\end{equation}

%当$\Omega_t = \Omega_g = \emptyset$的时候，说明该区间没有被历史或当前观测覆盖，因此直接舍去该区间。$\Omega_t \neq \emptyset \land \Omega_g \neq \emptyset$时，说明我们获得了一致的观测，因此使用贝叶斯滤波器在 p_j^g的基础上增加静态概率：
When \( \Omega_t = \Omega_g = \emptyset \), this means that the interval is not covered by any historical or current observations, thus we discard this interval. When \( \Omega_t \neq \emptyset \land \Omega_g \neq \emptyset \), it indicates that we have consistent observations, so we apply the Bayesian filter to increase the occupancy probability based on \( p_j^g \):
\begin{equation}
        p_k^{new} = \mathrm{Bayesfilter}(p_j^g, \alpha, \beta)
        \label{eq:confirmed_update}
\end{equation}

%当$\Omega_t = \emptyset \land \Omega_g \neq \emptyset$说明该区间仅获得了历史观测，而当前扫描未能获得该区间的观测。在这种情况下，该区间的静态概率将基于p_j^g进行降低：
When \( \Omega_t = \emptyset \land \Omega_g \neq \emptyset \), the height interval exists in the global pillar but remains unobserved in the current scan \( S_t \). Consequently, its occupancy probability is reduced based on \( p_j^g \):
\begin{equation}
        p_k^{new} = \mathrm{Bayesfilter}(p_j^g, 1-\alpha, 1-\beta)
        \label{eq:negative_update}
\end{equation}

%$当\Omega_t \neq \emptyset \land \Omega_g = \emptyset$我们基于$p_{empty} &更新概率，作为该高度区间的静态概率：
When \( \Omega_t \neq \emptyset \land \Omega_g = \emptyset \), we update the probability based on \( p_{\text{empty}} \) as the static probability for the new height interval:
\begin{equation}
p_k^{\text{new}} = \mathrm{Bayesfilter}(p_{\text{empty}}, \alpha, \beta)
\label{eq:novel_update}
\end{equation}

\begin{table*}[tp]
        \vspace{0.4cm}
        \caption{Accuracy Performance of Different Algorithms on Various Sequences}

        \label{table_acc}
        \centering
        \vspace{0.4cm}
        \begin{tabular}{l|c c c|c c c|c c c|c c c }
                \toprule
                & \multicolumn{3}{|c|}{KITTI~00}  & \multicolumn{3}{|c|}{KITTI~02} & \multicolumn{3}{|c|}{KITTI~05}  & \multicolumn{3}{|c }{AV2}\\
                \midrule
                Methods & SA(\%) & DA(\%) & \cellcolor{gray!10}AA(\%) & SA(\%) & DA(\%) & \cellcolor{gray!10}AA(\%) & SA(\%) & DA(\%) & \cellcolor{gray!10}AA(\%) & SA(\%) & DA(\%) & \cellcolor{gray!10}AA(\%) \\
                \midrule
                Removert\cite{kim2020remove} & 99.56 & 37.33 & \cellcolor{gray!10}60.97 & 96.67 & 39.08 & \cellcolor{gray!10}61.47 & 99.48 & 26.52 & \cellcolor{gray!10}51.37 & 98.97 & 31.16 & \cellcolor{gray!10}55.53 \\
                Erasor\cite{lim2021erasor} & 70.41 & 98.26 & \cellcolor{gray!10}83.18 & 50.76 & 98.60 & \cellcolor{gray!10}70.74 & 70.79 & 98.75 & \cellcolor{gray!10}83.61 & 77.51 & 99.18 & \cellcolor{gray!10}87.68 \\
                Octomap\cite{hornung2013octomap} & 69.57 & 99.71 & \cellcolor{gray!10}83.29 & 62.61 & 98.55 & \cellcolor{gray!10}78.55 & 69.37 & 98.34 & \cellcolor{gray!10}82.60 & 69.04 & 97.50 & \cellcolor{gray!10}82.04 \\
                Dynablox\cite{schmid2023dynablox} & 99.72 & 89.47 & \cellcolor{gray!10}94.46 & 83.07 & 94.86 & \cellcolor{gray!10}\underline{88.77} & 99.34 & 83.34 & \cellcolor{gray!10}90.99 & 99.35 & 86.08 & \cellcolor{gray!10}\underline{92.48} \\
                Dufomap\cite{duberg2024dufomap} & 98.64 & 98.53 & \cellcolor{gray!10}\textbf{98.58} & 68.56 & 98.47 &\cellcolor{gray!10} 82.16 & 97.18 & 95.44 & \cellcolor{gray!10}\textbf{96.31} & 96.67 & 88.90 & \cellcolor{gray!10}\textbf{92.70} \\
                HIF(Ours) & 98.40 & 95.46 & \cellcolor{gray!10}\underline{96.92} & 97.66 & 85.62 & \cellcolor{gray!10}\textbf{91.44} & 98.11 & 90.23 & \cellcolor{gray!10}\underline{94.09} & 83.47 & 84.17 & \cellcolor{gray!10}83.82 \\
                \bottomrule
        \end{tabular}
        \caption*{Note: \textbf{Bold} indicates the best performance in each column, \underline{Underlined} values denote the second-best performance}
\end{table*}
\subsection{Low-Height Preservation Strategy}
\label{LHP_S}
In our approach, we propose the Low-Height Preservation (LHP) strategy, which leverages the spatial characteristics of the pillar-based structure to prevent incorrect updates of unknown space in a new scan. Specifically, when matching indices between \( P_{mn}^t \) and \( P_{mn}^g \), we traverse each height interval in \( P_{mn}^t \), denoted as \( h_i^t = (b_i^t, t_i^t, p_i^t) \) for \( i = 1, 2, \dots, I \). We then extract the minimum \( b^t \) and define it as the base height \( b_{\text{base}} \) of the current pillar.

This \( b_{\text{base}} \) has two possible physical interpretations:
\begin{enumerate}
    \item It may correspond to the ground level observed within the pillar.
    \item Alternatively, it could be the lowest detected point, constrained by occlusion from objects in the foreground.
\end{enumerate}

The region below \( b_{\text{base}} \) is treated as unknown space in the current scan.

%%在我们的方法中，我们引入了低高度保持策略（Low-Height Preservation，LHP），巧妙地利用了pillar基础结构的空间特性，从而避免了新扫描中未知空间的错误更新。具体而言，当通过索引匹配得到 \( P_{mn}^t \) 和 \( P_{mn}^g \) 时，我们遍历 \( P_{mn}^t \) 中的每个高度区间，表示为 \( h_i^t = (b_i^t, t_i^t, p_i^t) \)，其中 \( i = 1, 2, \dots, I \)。我们提取出最小的 \( b^t \) 值，并将其视为当前pillar的**基础最低点** \( b_{\text{base}} \)。该基础最低点具有两种可能的物理含义：一方面，它可能代表当前pillar中观测到的地面；另一方面，它也可能是由于前方物体遮挡所观察到的最低点。低于该最低点的空间被认为是当前扫描中的未知空间。

%%为了定义这一最低点，我们通过以下公式从 \( P_{mn}^t \) 中的每个高度区间提取最小的 \( b^t \) 值作为基础最低点：
To define the lowest base point, we extract the minimum \( b^t \) value from all height intervals in \( P_{mn}^t \), using the following formula:  
\begin{equation}
b_{\text{base}} = \min\{ b_i^t \mid i = 1, 2, \dots, I \}
\end{equation}

%%在更新全局高度区间时，我们应用低高度保持策略，保持该区间的静态概率为：
When updating the global height intervals, we apply the Low-Height Preservation (LHP) strategy, maintaining the occupancy probability for the interval as:  

\begin{equation}
        p_k^{\text{new}} = p_j^g \quad \text{if} \quad \Omega_t = \emptyset \land \Omega_g \neq \emptyset, \, t_k \leq b_{\text{base}}
        \end{equation}

%%为了防止概率过度收敛，我们应用限幅更新：
To prevent excessive probability convergence, we apply a bounded update:
\begin{equation}
p_h^{\text{new}} = \text{clip}(p_h^{\text{new}}) \triangleq \max(0.1, \min(0.9, p_h^{\text{new}}))
\end{equation}

%%随后，我们用更新后的高度区间集合 \( \mathcal{H}^{\text{new}} \) 替换全局pillar \( P_{mn}^g \) 中的高度区间集合：
Finally, we replace the height intervals in the global pillar \( P_{mn}^g \) with the updated set \( \mathcal{H}^{\text{new}} \):  
\begin{equation}
P_{mn}^g \leftarrow \mathcal{H}^{\text{new}} \circ \text{clip}(p_h^{\text{new}})
\end{equation}
where \( \circ \) denotes the element-wise probability clipping operation.

%%其中，\( \circ \) 表示逐元素的概率限幅操作。

\section{Experiment}
Our experiments include Experimental Setup for method selection and parameters, Evaluation Datasets for dataset configurations, Evaluation Metrics for assessment criteria, Accuracy and Runtime Evaluation for performance analysis, and Ablation Study for key parameter impact.

\subsection{Experimental Setup}
We classify the compared methods into online and offline approaches, selecting representative works from each category. For offline methods, we include Removert, ERASOR, and the latest SOTA method, DUFOMap. For online methods, we evaluate the classic OctoMap and Dynablox. 

For parameter settings, we use the benchmark configurations from \cite{zhang2023dynamic} for Removert, ERASOR, and Dynablox. OctoMap is set with a voxel size of 0.1 meters. For DUFOMap, we follow the original paper’s parameters, using a voxel size of 0.1 meters, $d_s = 0.2$ meters, and a sub-voxel localization error of $d_p = 1$.

%%我们将工作分为在线方法和离线方法进行比较，并分别选取了具有代表性的工作。对于离线方法，我们选择了Removert、ERASOR和最新的SOTA方法DUFOMap；对于在线方法，我们选择了经典的OctoMap和Dynablox进行比较。对于Removert、ERASOR和Dynablox，我们使用了benchmark \cite{zhang2023dynamic}中的参数进行评估。OctoMap的设置为体素大小0.1米，而DUFOMap则参照原论文中的参数，其中体素大小为0.1米，ds = 0.2米，子体素定位误差dp = 1。

For our method, we set the pillar size to 1 meter along both the X and Y axes. For different datasets, we only fine-tuned the observation model parameters $P_S$ and $P_D$.

%%对于我们的方法，我们设置pillar在XY方向上均为边长1米，且对于不同的数据集，仅对观测模型参数$P_S$和$P_D$进行了微调。

All experiments were conducted on a desktop computer equipped with an Intel Core i9-13900KF processor and 64GB of RAM.
%%所有实验均在一台配备Intel Core i9-13900KF处理器和64GB内存的桌面计算机上进行。

\subsection{Evaluation Datasets}

To validate the effectiveness of our algorithm, we conducted a precision evaluation and runtime evaluation using the \textbf{KITTI} dataset, with ground truth labels provided by \textbf{Semantic KITTI}. The dataset is captured using the HDL-64E LiDAR and includes pose and point cloud labels. Following the setup in \cite{lim2021erasor} and \cite{zhang2023dynamic}, we selected \textbf{sequences 00} (frames \textbf{4,390 - 4,530}), \textbf{01} (frames \textbf{150 - 250}), \textbf{02} (frames \textbf{860 - 950}), and \textbf{05} (frames \textbf{2,350 - 2,670}). During preprocessing, we discarded points labeled as \textbf{0 (unknown)} and \textbf{1 (unlabeled)}.

%%为了验证算法的有效性，我们在使用HDL-64E LiDAR的Semantic KITTI数据集上进行了精度评估，并通过序列运行时间评估其性能。该数据集提供了位姿和点云标签信息。遵循\cite{lim2021erasor}和\cite{zhang2023dynamic}的工作，我们选择了序列00（帧4,390 - 4,530）、01（帧150 - 250）、02（帧860 - 950）和05（帧2,350 - 2,670），并在预处理阶段去除了标签为0（未知）和1（未标记）的点。

To evaluate the adaptability of our algorithm to different sensors, we conducted supplementary experiments on the \textbf{Argoverse 2} dataset, which utilizes the \textbf{VLP-32C LiDAR}. The dataset setup follows the configuration in \cite{zhang2023dynamic}.

%%为了验证算法在不同传感器上的适用性，我们还使用了VLP-32C LiDAR的Argoverse 2数据集进行补充实验。数据集设置参照\cite{zhang2023dynamic}。

\begin{figure*}[htbp]  % 't' 表示置顶
        \centering
        \vspace{0.3cm}
        \includegraphics[width=\textwidth]{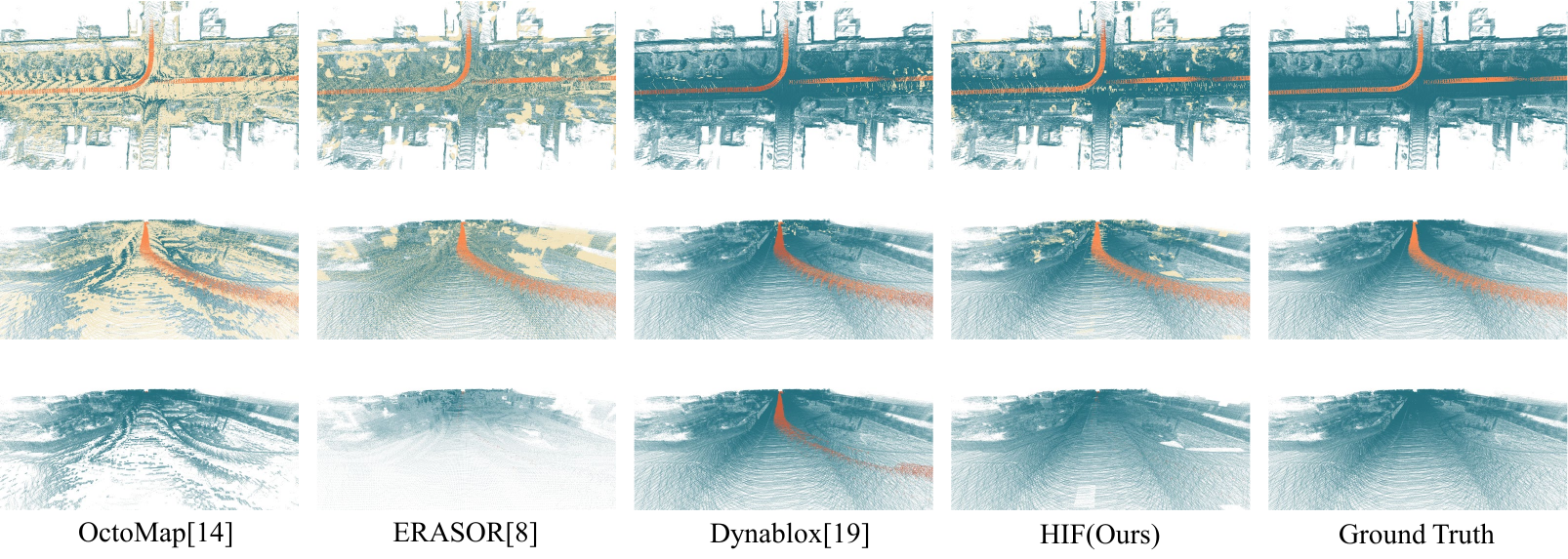}  % 这里替换为您的图片路径
        \caption{Comparison of different algorithms on the KITTI~05 sequence. The first two rows show the classification results for all points by each algorithm, while the last row presents the final output point clouds. Colors indicate classification outcomes: \textcolor[rgb]{0.145, 0.443, 0.502}{(blue)} for correctly retained static points, \textcolor[rgb]{0.992, 0.545, 0.318}{(orange)} for correctly removed dynamic points, \textcolor[rgb]{0.949, 0.898, 0.749}{(yellow)} for misclassified static points, and \textcolor[rgb]{0.796, 0.376, 0.251}{(red)} for missed dynamic points.}
        %图示展示了不同算法在KITTI05序列上的表现。前两行展示了各算法对所有点的分类结果，最后一行展示了算法输出的点云。在图中，点的颜色表示如下：正确保留的静态点用\textcolor[rgb]{0.145, 0.443, 0.502}{(蓝色)}表示；正确去除的动态点用\textcolor[rgb]{0.992, 0.545, 0.318}{(橙色)}表示；误分类的静态点用\textcolor[rgb]{0.949, 0.898, 0.749}{(黄色)}表示；漏检测的动态点用\textcolor[rgb]{0.796, 0.376, 0.251}{(红色)}表示。
        \label{algo_comp}  % 图片的引用标签
\end{figure*}

\begin{table*}[htbp]
        \caption{Runtime Performance of Different Algorithms on the KITTI Dataset}
        \label{table_time_example}
        \centering
        \resizebox{\textwidth}{!}{
                \begin{tabular}{l|c c c|c c c|c c c|c c c|c c c}
                        \toprule
                         & \multicolumn{3}{c}{KITTI~00}  & \multicolumn{3}{|c}{KITTI~01} & \multicolumn{3}{|c}{KITTI~02}  & \multicolumn{3}{|c}{KITTI~05}   & \multicolumn{3}{|c}{KITTI~07}\\
                        \midrule
                        Methods & Mean & Std & FPS & Mean & Std & FPS & Mean & Std & FPS & Mean & Std & FPS & Mean & Std & FPS \\
                        \midrule
                        Removert\cite{kim2020remove} & \underline{56.74} & \underline{1.149} & \underline{17.62} & \underline{39.07} & \textbf{0.925} & \underline{25.60} & \underline{38.07} & \textbf{0.49} & \underline{26.27} & 127.7 & \underline{2.615} & 7.831 & 76.87 & \textbf{1.517} & 13.01 \\
                        Erasor\cite{lim2021erasor} & 105.3 & 13.31 & 9.497 & 708.2 & 40.35 & 1.412 & 98.20 & 24.32 & 10.18 & 203.9 & 19.69 & 4.904 & 57.52 & 12.33 & \underline{17.39} \\
                        Octomap\cite{hornung2013octomap} & 887.5 & 92.95 & 1.126 & 2225 & 665.4 & 0.494 & 733.5 & 73.71 & 1.363 & 1077 & 121.7 & 0.929 & 792.4 & 71.85 & 1.262\\
                        Dynablox\cite{schmid2023dynablox} & 79.13 & 13.41 & 12.64 & 180.7 & 49.52 & 5.534 & 81.85 & 17.88 & 12.22 & \underline{84.49} & 14.81 & \underline{11.84} & \underline{74.10} & 10.85 & 13.50 \\
                        Dufomap\cite{duberg2024dufomap} & 98.01 & 16.85 & 10.20 & 127.7 & 21.10 & 7.831 & 83.80 & 17.68 & 11.93 & 93.46 & 19.61 & 10.70 & 81.20 & 20.03 & 12.32 \\
                        HIF(Ours) & \textbf{11.62} & \textbf{0.878} & \textbf{86.06} & \textbf{12.05} & \underline{1.784} & \textbf{82.99} & \textbf{11.70} & \underline{1.310} & \textbf{90.33} & \textbf{13.09} & \textbf{1.809} & \textbf{76.39} & \textbf{13.32} & \underline{1.957} & \textbf{75.08} \\
                        \bottomrule
                \end{tabular}
        }
        \caption*{Note: Mean denotes the average runtime, Std represents the standard deviation of the runtime, FPS indicates the algorithm's frame rate. The units for Mean and Std are ms, and the unit for FPS is Hz. \textbf{Bold} indicates the best performance in each column, \underline{Underlined} values denote the second-best performance}
        
\end{table*}

\subsection{Evaluation Metrics}

%我们遵循\cite{zhang2023dynamic}中的评价方式，直接使用点级别的静态精度（SA）和动态精度（DA）来更直接地反映算法的准确性，而不是通过对地面真实值进行下采样来获得评估结果。同时，我们采用了关联精度（AA）作为综合评价指标，定义如公式\ref{eq:Associated_accuracy}所示，以确保对静态点和动态点的准确性进行综合评估。

We follow the evaluation approach of \cite{zhang2023dynamic}, using point-level \textbf{static accuracy (SA)} and \textbf{dynamic accuracy (DA)} to directly measure the algorithm's performance, avoiding the need for ground truth downsampling. Additionally, we employ \textbf{Associated Accuracy (AA)} as a comprehensive metric, defined in Eq.~\ref{eq:Associated_accuracy}, to ensure sensitivity to both static and dynamic points:
\begin{equation}
AA = \sqrt{SA \times DA}
\label{eq:Associated_accuracy}
\end{equation}

For runtime evaluation, we measure the average runtime per frame and FPS to assess computational efficiency, while the standard deviation of runtime is used to evaluate the algorithm's stability.

%我们还使用平均每帧运行时间和FPS作为算法运行速度的评判指标，并关注运行时间的标准差来评判算法的稳定性。

\subsection{Accuracy Evaluation}
We evaluated our method on the \textbf{KITTI} and \textbf{AV2} datasets, with results presented in Table~\ref{table_acc}. \textbf{Removert} achieves high \textbf{static accuracy (SA)} but struggles with \textbf{dynamic accuracy (DA)}, indicating difficulty in detecting dynamic objects, whereas \textbf{ERASOR} and \textbf{OctoMap} perform well in dynamic object detection but suffer from significant misclassification, leading to lower static accuracy. \textbf{Dynablox}, \textbf{DUFOMap}, and our method \textbf{HIF} maintain a good balance between static and dynamic accuracy, with \textbf{HIF} ranking just below the SOTA method \textbf{DUFOMap} on \textbf{KITTI} while achieving the \textbf{best performance on KITTI~02}, and on \textbf{AV2}, the primary gap between \textbf{HIF} and \textbf{DUFOMap} lies in static accuracy, which we further analyze in the following visualization.

% 我们在KITTI和AV2数据集上进行了实验评估，如表\ref{table_acc}所示。Removert方法在静态精度上表现优异，但在动态精度上略显不足，表明其在有效检测动态物体时存在困难。ERASOR和OctoMap在动态物体检测方面表现出色，但都存在较为严重的误杀现象，导致静态精度较低。Dynablox、DUFOMap和我们的方法HIF在动态精度和静态精度之间达到了较好的平衡。在KITTI数据集上，我们的方法仅次于SOTA方法DUFOMap，并且在02序列上表现出了较大的优势，取得了最佳性能。而在AV2数据集上，我们的方法与SOTA方法DUFOMap的差距主要体现在静态精度，接下来的可视化分析将进一步探讨这一原因。

\begin{figure*}[htbp]
        \centering
        \vspace{0.2cm}
        \includegraphics[width=\textwidth]{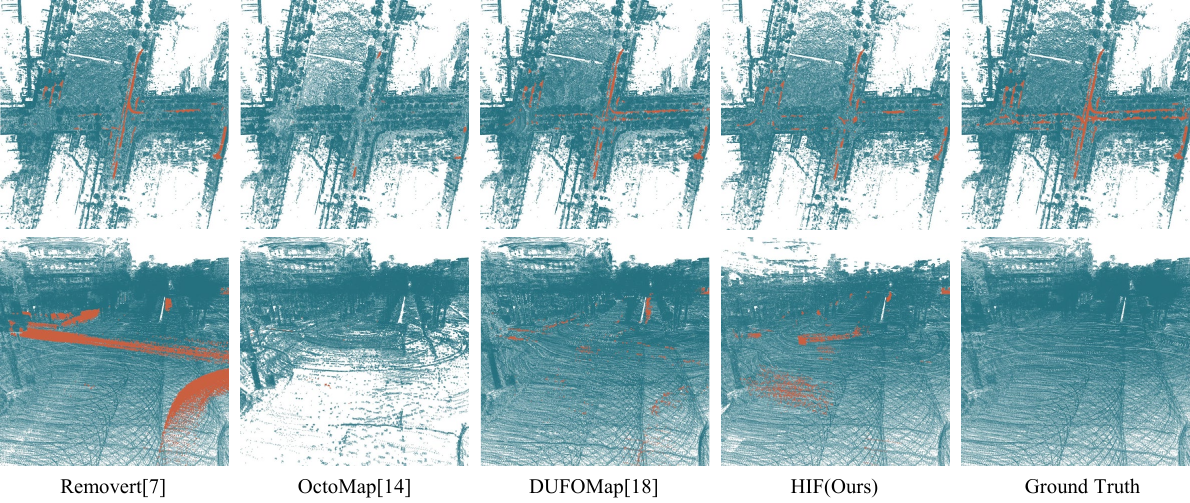}  % 这里替换为您的图片路径
        \caption{Comparison of different algorithms on the AV2 sequence. Colors indicate classification outcomes: \textcolor[rgb]{0.145, 0.443, 0.502}{(blue)} for correctly retained static points and \textcolor[rgb]{0.796, 0.376, 0.251}{(red)} for missed dynamic points.}
        %图示展示了不同算法在AV2序列上的输出结果。图中，点的颜色表示如下：正确保留的静态点用\textcolor[rgb]{0.145, 0.443, 0.502}{(蓝色)}表示；漏检测的动态点用\textcolor[rgb]{0.796, 0.376, 0.251}{(红色)}表示。
        \label{fig:av2_comp}  % 图片的引用标签
\end{figure*}
First, we selected the representative \textbf{KITTI~05} sequence from the \textbf{KITTI} dataset for visualization, as shown in Fig.~\ref{algo_comp}. \textbf{OctoMap} and \textbf{ERASOR} exhibit significant misclassification issues: the former frequently misclassifies ground points, while the latter suffers from sparse ground point retention due to insufficient preservation in the segmentation module, leading to entire pillars being misclassified. \textbf{Dynablox} excels in retaining static points but struggles with detecting dynamic objects. In contrast, our method, \textbf{HIF}, achieves strong performance for both static and dynamic points, with only minor misclassification observed at the top of the canopy.

Then, as shown in Fig.~\ref{fig:av2_comp}, \textbf{Removert} effectively retains static points but struggles with dynamic object removal. Compared to the SOTA method \textbf{DUFOMap}, our approach shows weaker performance in high-rise buildings and canopy areas, with more frequent misclassifications. This is primarily due to the fact that in actual scan data, higher regions appear less frequently within their respective pillars due to occlusions or radar scanning angle limitations, leading to an incorrect reduction in static probabilities.

%在图\ref{fig:av2_comp}中可以看到，Removert能够很好保留静态点，但在动态点去除方面存在不足。与SOTA方法DUFOMap相比，我们的方法主要在高处建筑和树冠部分表现较差，误杀现象较多。这是由于在实际扫描数据中，由于遮挡或雷达扫描角度的限制，导致高处部分在其所属pillar中的出现频率较低，因而被错误的降低静态概率。

\subsection{Runtime Evaluation}
We evaluated the runtime of all methods on the \textbf{KITTI} sequences, with the results presented in Table~\ref{table_time_example}. Our method demonstrated a significant advantage across all sequences, achieving a speedup of several times compared to other approaches. Notably, its runtime remains consistent across different sequences, indicating stable performance across varying environments, whether open or narrow. Furthermore, the per-frame runtime variation is minimal, with low variance, highlighting the method's robustness. The combination of fast execution and high stability makes our algorithm well-suited for integration into existing SLAM systems, enhancing performance without introducing instability.

%我们在KITTI数据集上对所有方法的运行时间进行了评估，结果如表\ref{table_time_example}所示。我们的方法在所有序列上都取得了显著的优势，最大场景下甚至能取得高达100倍的速度提升。值得注意的是，我们的方法的运行时间在整个数据集上保持一致，这表明算法在不同的场景下都能表现出良好的一致性，无论环境是开阔还是狭窄。此外，每帧的运行时间的方差较低，这意味着算法非常稳定。快速的运行时间和稳定性为我们算法的集成到现有的SLAM系统中提供了保障。

\begin{figure}[thpb]
        \centering
        \resizebox{\linewidth}{!}{%
                \includegraphics[height=2cm]{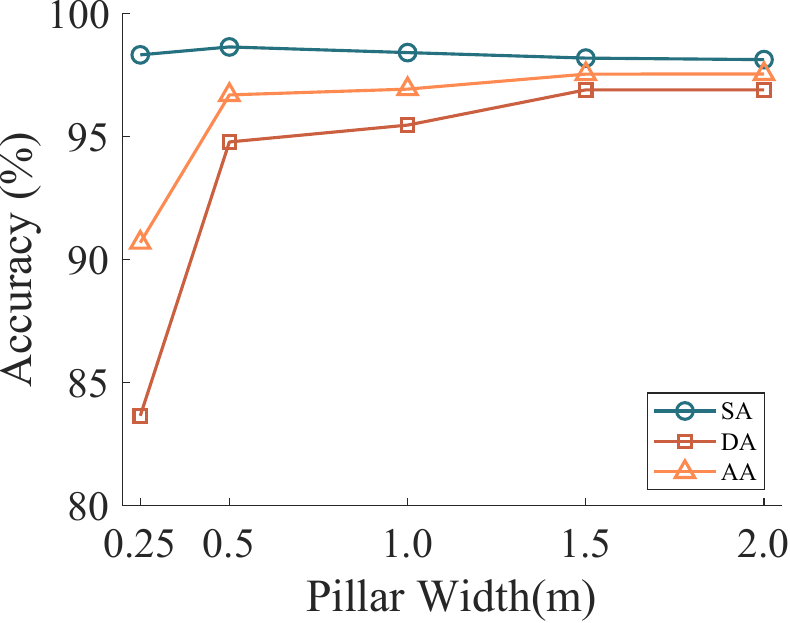}%
                \hspace{0.01\linewidth} % 图片间隔
                \includegraphics[height=2cm]{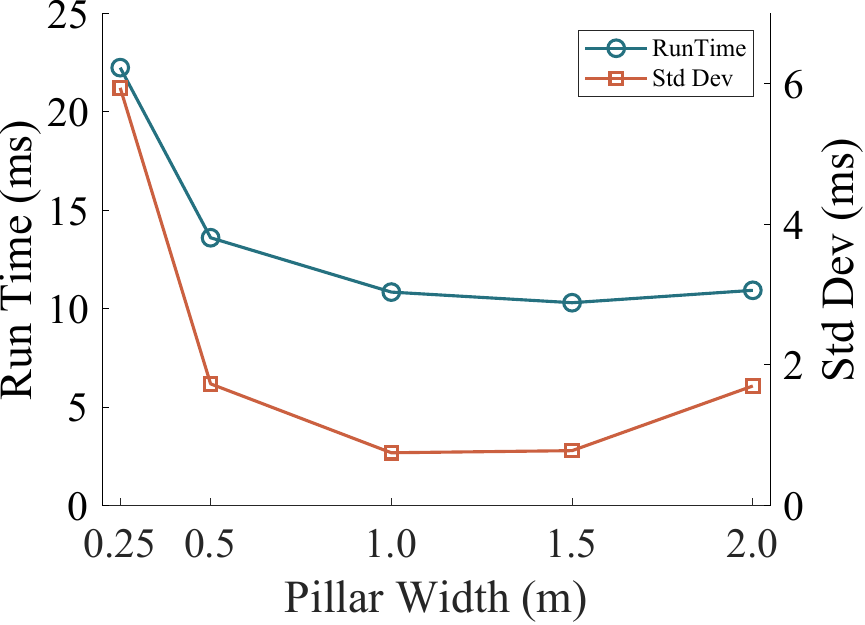}
        }
        \caption{Impact of Pillar Width on the Method's Performance}

        \label{fig:combined}
\end{figure}

To analyze the influence of pillar width on both accuracy and computational efficiency, we conducted a series of experiments (Fig.~\ref{fig:combined}). When the pillar width is small, a high miss rate is observed in dynamic object detection. As the width increases, dynamic accuracy (DA) improves and converges to 95.5\%, while static accuracy (SA) remains nearly unchanged.

\begin{figure}[thpb]
        \centering
        \includegraphics[width=0.8\linewidth]{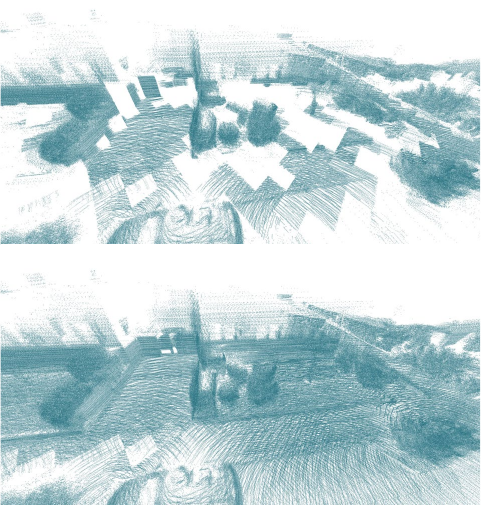}

        \caption{Ablation study of the LHP module. The top image shows the output without using the LHP module, while the bottom image shows the output with the LHP module.}
        %图示展示了LHP模块的消融实验。上图展示了未使用LHP模块的算法输出，下图展示了使用LHP模块的算法输出。
        \label{fig:LHP}
\end{figure}
        
Furthermore, when the pillar width is small, the algorithm exhibits higher runtime and greater variance. In contrast, at 1 meter and 1.5 meters, the algorithm achieves the best runtime performance.

Additionally, we found that regardless of the pillar width, the algorithm’s peak memory usage remained stable at 801 ± 1MB, indicating high memory efficiency. The maximum memory consumption is primarily determined by the scale of the point cloud in the scene rather than the pillar width.

% The ablation results on pillar width are shown in Fig.~\ref{fig:combined}. When the pillar width is small, dynamic object detection suffers from a high miss rate. As the width increases, \textbf{dynamic accuracy (DA)} improves and stabilizes around 95.5\%, while \textbf{static accuracy (SA)} experiences a slight decline.

% % 我们对pillar width进行消融实验的结果如图\ref{fig:1}和图\ref{fig:2}所示。可以看出，当pillar width较小时，动态物体的漏检问题较为严重；随着pillar width的增大，动态准确度（DA）有所提升，并趋于稳定在95.5%左右。同时，静态准确度（SA）随着pillar width增大略有下降。

% Moreover, when the pillar width is set to \textbf{0.25 meters}, both the per-frame runtime and its variance are relatively high. In contrast, with a pillar width of \textbf{1 meter} or \textbf{1.5 meters}, the runtime is lower and more stable. Considering these factors, the optimal pillar width for our algorithm on the \textbf{KITTI~00} sequence is approximately \textbf{1 meter}. Additionally, we observed that regardless of the pillar width, the algorithm's maximum memory usage remained stable at \textbf{801±1MB}, indicating high memory efficiency, with peak memory consumption dependent solely on the scale of the point cloud in the scene.

% % 进一步地，正如图\ref{fig:2}所示，当pillar宽度为0.25米时，单帧的运行时间和运行方差较大，而当pillar宽度为1米或1.5米时，运行时间较低且较为稳定。因此，综合考虑，我们算法在KITTI00序列上的最佳pillar width取值约为1米。同时，在实验中我们发现，无论pillar宽度如何设置，算法的最大内存使用始终维持在801±1MB，这表明我们的算法具有较高的内存效率，最大内存量仅取决于场景中点云的规模。

\subsection{Ablation Study}

Since Height Interval Construction and Interval Probability Filtering form the core components of our approach, we conduct ablation experiments exclusively on the Low-Height Preservation (LHP) module. To evaluate its effectiveness, we test this module across all data sequences.

%%我们对主要参数pillar width和LHP模块进行了消融实验。具体而言，我们在KITTI00序列上对pillar width参数进行了精准度和运行时间的综合评估，并在所有序列中测试了LHP模块的有效性。

\begin{table}[thpb]
        \centering
        \caption{Ablation Study of the LHP Module}

        \label{tab:ablation_results}
        \begin{tabularx}{0.7\columnwidth}{>{\centering\arraybackslash}X >{\centering\arraybackslash}X | *3{>{\centering\arraybackslash}X}} % 设置总宽度为单栏的80%
            \toprule Seq & LHP & SA(\%) & DA(\%) & AA(\%) \\
            \midrule
            \multirow{2}{*}{00} & w/o  & 91.00 & 95.48 & 93.22 \\
                                & w/   & 98.40 & 95.46 & 96.92  \\
            \midrule
            \multirow{2}{*}{01} & w/o  & 97.78 & 65.23 & 79.86 \\
                                & w/   & 99.20 & 63.82 & 79.57 \\
            \midrule
            \multirow{2}{*}{02} & w/o  & 64.25 & 90.68 & 76.33 \\
                                & w/   & 97.66 & 85.62 & 91.44 \\
            \midrule
            \multirow{2}{*}{05} & w/o  & 94.00 & 90.52 & 92.24 \\
                                & w/   & 98.11 & 90.23 & 94.09 \\
            \midrule
            \multirow{2}{*}{07} & w/o  & 92.84 & 68.92 & 79.99 \\
                                & w/   & 99.00 & 65.38 & 80.46 \\
            \midrule
            \multirow{2}{*}{AV2} & w/o  & 74.08 & 85.23 & 79.46 \\
                                & w/   & 83.47 & 84.17 & 83.82 \\
            \bottomrule
        \end{tabularx}
\end{table}
    
Results presented in Table~\ref{tab:ablation_results}, showing that while \textbf{LHP} has a negligible impact on \textbf{dynamic accuracy (DA)}, it significantly improves \textbf{static accuracy (SA)}; to further investigate this effect, we visualized the results for the \textbf{KITTI~02} sequence in Fig.~\ref{fig:LHP}, where the complex scene—featuring parked vehicles, bushes, and occluded objects—leads to relatively low static accuracy for most algorithms, as shown in Table~\ref{table_acc}, whereas our \textbf{LHP} strategy effectively identifies these occlusions and preserves static points in occluded areas.

%%我们在所有序列中测试了LHP策略的有效性，结果如表\ref{tab:ablation_results}所示。可以看出，在对动态准确度影响可以忽略的情况下，LHP策略能够有效提升静态准确度（SA），从而提高综合指标AA。为进一步分析这一现象的原因，我们选取了KITTI~02序列进行可视化，如图\ref{fig:LHP}所示。图中显示，KITTI~02序列的场景较为复杂，存在停放的车辆和灌木等物体，这些物体在激光雷达扫描过程中会从不同角度遮挡后方的物体；同时，场景中许多较深的角落在每帧扫描中并不能完全被观测到。这些复杂的空间关系导致大多数算法在该序列上的静态准确度（SA）较差，如表\ref{table_acc}所示。而我们的LHP策略能够有效识别这些遮挡关系，并保持被遮挡区域的静态点。

\section{Conclusions}

In this paper, we introduced \textbf{HIF}, a novel dynamic object removal framework that models occupancy probabilities along the vertical direction. Our method constructs \textbf{pillar-based height interval} representations, enabling efficient spatial partitioning, and employs Bayesian filtering to iteratively update interval probabilities. Additionally, the proposed \textbf{low-height preservation (LHP)} strategy effectively detects unknown spaces, improving robustness in highly occluded environments.

Experimental results demonstrate that \textbf{HIF} achieves state-of-the-art runtime efficiency while maintaining high accuracy, outperforming conventional methods in both computational speed and stability. However, certain limitations remain: misclassification can occur when high-altitude regions are observed less frequently than their corresponding pillars, and dynamic objects may be missed in areas with low observation frequency. Future work will focus on refining the pillar representation and exploring adaptive filtering techniques to further improve accuracy and robustness in diverse real-world scenarios.

%%本文提出了HIF算法，通过构建高度区间对垂直信息进行概率建模，并且使用贝叶斯滤波器更新区间概率，实现了在线实时点云动态物体去除。此外，低高度保持策略有效识别未知空间，提高了算法在复杂场景下的变现。与现有方法相比，HIF实现了最快的运行速度和较高的稳定性。然而，当前算法仍存在一些问题：当高处区域的柱体观察频率较高而高处区间的观察频率较低时，可能会导致误杀现象；而对于低频观测区域，动态物体可能会漏检。未来的研究将致力于改进这些问题，进一步提升算法的鲁棒性和精度。

\addtolength{\textheight}{-12cm}    % This command serves to balance the column lengths

%%%%%%%%%%%%%%%%%%%%%%%%%%%%%%%%%%%%%%%%%%%%%%%%%%%%%%%%%%%%%%%%%%%%%%%%%%%%%%%%

\bibliographystyle{IEEEtran}  % 选择参考文献样式，如IEEE格式
\bibliography{references}     % 引入 references.bib 文件，不需要 .bib 后缀 

\begin{thebibliography}{10}
\providecommand{\url}[1]{#1}
\csname url@samestyle\endcsname
\providecommand{\newblock}{\relax}
\providecommand{\bibinfo}[2]{#2}
\providecommand{\BIBentrySTDinterwordspacing}{\spaceskip=0pt\relax}
\providecommand{\BIBentryALTinterwordstretchfactor}{4}
\providecommand{\BIBentryALTinterwordspacing}{\spaceskip=\fontdimen2\font plus
\BIBentryALTinterwordstretchfactor\fontdimen3\font minus \fontdimen4\font\relax}
\providecommand{\BIBforeignlanguage}[2]{{%
\expandafter\ifx\csname l@#1\endcsname\relax
\typeout{** WARNING: IEEEtran.bst: No hyphenation pattern has been}%
\typeout{** loaded for the language `#1'. Using the pattern for}%
\typeout{** the default language instead.}%
\else
\language=\csname l@#1\endcsname
\fi
#2}}
\providecommand{\BIBdecl}{\relax}
\BIBdecl

\bibitem{shan2020lio}
T.~Shan, B.~Englot, D.~Meyers, W.~Wang, C.~Ratti, and D.~Rus, ``Lio-sam: Tightly-coupled lidar inertial odometry via smoothing and mapping,'' in \emph{2020 IEEE/RSJ international conference on intelligent robots and systems (IROS)}.\hskip 1em plus 0.5em minus 0.4em\relax IEEE, 2020, pp. 5135--5142.

\bibitem{xu2022fast}
W.~Xu, Y.~Cai, D.~He, J.~Lin, and F.~Zhang, ``Fast-lio2: Fast direct lidar-inertial odometry,'' \emph{IEEE Transactions on Robotics}, vol.~38, no.~4, pp. 2053--2073, 2022.

\bibitem{yoon2019mapless}
D.~Yoon, T.~Tang, and T.~Barfoot, ``Mapless online detection of dynamic objects in 3d lidar,'' in \emph{2019 16th Conference on Computer and Robot Vision (CRV)}.\hskip 1em plus 0.5em minus 0.4em\relax IEEE, 2019, pp. 113--120.

\bibitem{pagad2020robust}
S.~Pagad, D.~Agarwal, S.~Narayanan, K.~Rangan, H.~Kim, and G.~Yalla, ``Robust method for removing dynamic objects from point clouds,'' in \emph{2020 IEEE International Conference on Robotics and Automation (ICRA)}.\hskip 1em plus 0.5em minus 0.4em\relax IEEE, 2020, pp. 10\,765--10\,771.

\bibitem{fu2022mapcleaner}
H.~Fu, H.~Xue, and G.~Xie, ``Mapcleaner: Efficiently removing moving objects from point cloud maps in autonomous driving scenarios,'' \emph{Remote Sensing}, vol.~14, no.~18, p. 4496, 2022.

\bibitem{wu2024otd}
R.~Wu, Z.~Fang, C.~Pang, and X.~Wu, ``Otd: an online dynamic traces removal method based on observation time difference,'' \emph{IEEE Transactions on Geoscience and Remote Sensing}, 2024.

\bibitem{kim2020remove}
G.~Kim and A.~Kim, ``Remove, then revert: Static point cloud map construction using multiresolution range images,'' in \emph{2020 IEEE/RSJ International Conference on Intelligent Robots and Systems (IROS)}.\hskip 1em plus 0.5em minus 0.4em\relax IEEE, 2020, pp. 10\,758--10\,765.

\bibitem{lim2021erasor}
H.~Lim, S.~Hwang, and H.~Myung, ``Erasor: Egocentric ratio of pseudo occupancy-based dynamic object removal for static 3d point cloud map building,'' \emph{IEEE Robotics and Automation Letters}, vol.~6, no.~2, pp. 2272--2279, 2021.

\bibitem{zhang2024erasor++}
J.~Zhang and Y.~Zhang, ``Erasor++: Height coding plus egocentric ratio based dynamic object removal for static point cloud mapping,'' \emph{arXiv preprint arXiv:2403.05019}, 2024.

\bibitem{jia2024beautymap}
M.~Jia, Q.~Zhang, B.~Yang, J.~Wu, M.~Liu, and P.~Jensfelt, ``Beautymap: Binary-encoded adaptable ground matrix for dynamic points removal in global maps,'' \emph{IEEE Robotics and Automation Letters}, 2024.

\bibitem{geiger2013vision}
A.~Geiger, P.~Lenz, C.~Stiller, and R.~Urtasun, ``Vision meets robotics: The kitti dataset,'' \emph{The international journal of robotics research}, vol.~32, no.~11, pp. 1231--1237, 2013.

\bibitem{behley2019semantickitti}
J.~Behley, M.~Garbade, A.~Milioto, J.~Quenzel, S.~Behnke, C.~Stachniss, and J.~Gall, ``Semantickitti: A dataset for semantic scene understanding of lidar sequences,'' in \emph{Proceedings of the IEEE/CVF international conference on computer vision}, 2019, pp. 9297--9307.

\bibitem{wilson2023argoverse}
B.~Wilson, W.~Qi, T.~Agarwal, J.~Lambert, J.~Singh, S.~Khandelwal, B.~Pan, R.~Kumar, A.~Hartnett, J.~K. Pontes \emph{et~al.}, ``Argoverse 2: Next generation datasets for self-driving perception and forecasting,'' \emph{arXiv preprint arXiv:2301.00493}, 2023.

\bibitem{hornung2013octomap}
A.~Hornung, K.~M. Wurm, M.~Bennewitz, C.~Stachniss, and W.~Burgard, ``Octomap: An efficient probabilistic 3d mapping framework based on octrees,'' \emph{Autonomous robots}, vol.~34, pp. 189--206, 2013.

\bibitem{schauer2018peopleremover}
J.~Schauer and A.~N{\"u}chter, ``The peopleremover—removing dynamic objects from 3-d point cloud data by traversing a voxel occupancy grid,'' \emph{IEEE robotics and automation letters}, vol.~3, no.~3, pp. 1679--1686, 2018.

\bibitem{zhang2023dynamic}
Q.~Zhang, D.~Duberg, R.~Geng, M.~Jia, L.~Wang, and P.~Jensfelt, ``A dynamic points removal benchmark in point cloud maps,'' in \emph{2023 IEEE 26th International Conference on Intelligent Transportation Systems (ITSC)}.\hskip 1em plus 0.5em minus 0.4em\relax IEEE, 2023, pp. 608--614.

\bibitem{arora2023static}
M.~Arora, L.~Wiesmann, X.~Chen, and C.~Stachniss, ``Static map generation from 3d lidar point clouds exploiting ground segmentation,'' \emph{Robotics and Autonomous Systems}, vol. 159, p. 104287, 2023.

\bibitem{duberg2024dufomap}
D.~Duberg, Q.~Zhang, M.~Jia, and P.~Jensfelt, ``Dufomap: Efficient dynamic awareness mapping,'' \emph{IEEE Robotics and Automation Letters}, 2024.

\bibitem{yan2023rh}
Z.~Yan, X.~Wu, Z.~Jian, B.~Lan, and X.~Wang, ``Rh-map: Online map construction framework of dynamic object removal based on 3d region-wise hash map structure,'' \emph{IEEE Robotics and Automation Letters}, 2023.

\bibitem{milioto2019rangenet++}
A.~Milioto, I.~Vizzo, J.~Behley, and C.~Stachniss, ``Rangenet++: Fast and accurate lidar semantic segmentation,'' in \emph{2019 IEEE/RSJ international conference on intelligent robots and systems (IROS)}.\hskip 1em plus 0.5em minus 0.4em\relax IEEE, 2019, pp. 4213--4220.

\bibitem{cortinhal2020salsanext}
T.~Cortinhal, G.~Tzelepis, and E.~Erdal~Aksoy, ``Salsanext: Fast, uncertainty-aware semantic segmentation of lidar point clouds,'' in \emph{Advances in Visual Computing: 15th International Symposium, ISVC 2020, San Diego, CA, USA, October 5--7, 2020, Proceedings, Part II 15}.\hskip 1em plus 0.5em minus 0.4em\relax Springer, 2020, pp. 207--222.

\bibitem{mersch2022receding}
B.~Mersch, X.~Chen, I.~Vizzo, L.~Nunes, J.~Behley, and C.~Stachniss, ``Receding moving object segmentation in 3d lidar data using sparse 4d convolutions,'' \emph{IEEE Robotics and Automation Letters}, vol.~7, no.~3, pp. 7503--7510, 2022.

\bibitem{chen2021moving}
X.~Chen, S.~Li, B.~Mersch, L.~Wiesmann, J.~Gall, J.~Behley, and C.~Stachniss, ``Moving object segmentation in 3d lidar data: A learning-based approach exploiting sequential data,'' \emph{IEEE Robotics and Automation Letters}, vol.~6, no.~4, pp. 6529--6536, 2021.

\bibitem{sun2022efficient}
J.~Sun, Y.~Dai, X.~Zhang, J.~Xu, R.~Ai, W.~Gu, and X.~Chen, ``Efficient spatial-temporal information fusion for lidar-based 3d moving object segmentation,'' in \emph{2022 IEEE/RSJ International Conference on Intelligent Robots and Systems (IROS)}.\hskip 1em plus 0.5em minus 0.4em\relax IEEE, 2022, pp. 11\,456--11\,463.

\bibitem{li2023efficient}
Q.~Li and Y.~Zhuang, ``An efficient image-guided-based 3d point cloud moving object segmentation with transformer-attention in autonomous driving,'' \emph{International Journal of Applied Earth Observation and Geoinformation}, vol. 123, p. 103488, 2023.

\bibitem{xie2023real}
X.~Xie, H.~Wei, and Y.~Yang, ``Real-time lidar point-cloud moving object segmentation for autonomous driving,'' \emph{Sensors}, vol.~23, no.~1, p. 547, 2023.

\bibitem{cheng2024mf}
J.~Cheng, K.~Zeng, Z.~Huang, X.~Tang, J.~Wu, C.~Zhang, X.~Chen, and R.~Fan, ``Mf-mos: A motion-focused model for moving object segmentation,'' \emph{arXiv preprint arXiv:2401.17023}, 2024.

\bibitem{zhang2022not}
Y.~Zhang, Q.~Hu, G.~Xu, Y.~Ma, J.~Wan, and Y.~Guo, ``Not all points are equal: Learning highly efficient point-based detectors for 3d lidar point clouds,'' in \emph{Proceedings of the IEEE/CVF conference on computer vision and pattern recognition}, 2022, pp. 18\,953--18\,962.

\bibitem{schmid2023dynablox}
L.~Schmid, O.~Andersson, A.~Sulser, P.~Pfreundschuh, and R.~Siegwart, ``Dynablox: Real-time detection of diverse dynamic objects in complex environments,'' \emph{IEEE Robotics and Automation Letters}, 2023.

\end{thebibliography}

\end{document}